\documentclass[sigconf,anonymous=false,natbib=true]{acmart}
\usepackage{graphicx}
\usepackage{amsmath}
\usepackage{subfigure}
\usepackage{subcaption}
\usepackage{makecell}
\usepackage{multirow}
\usepackage{algorithm}
\usepackage{algpseudocode}

\AtBeginDocument{%
  }

\setcopyright{acmlicensed}
\copyrightyear{2018}
\acmYear{2018}
\acmDOI{XXXXXXX.XXXXXXX}

\acmConference[Conference acronym 'XX]{Make sure to enter the correct
  conference title from your rights confirmation emai}{June 03--05,
  2018}{Woodstock, NY}
\acmISBN{978-1-4503-XXXX-X/18/06}

\begin{document}

\title{Application Specific Compression of Deep Learning Models}

\author{Rohit Raj Rai}
\email{rohitraj@iitg.ac.in}
\affiliation{%
  \institution{Indian Institute of Technology Guwahati \country{India}}
  \city{}
  \state{}
  }

\author{Angana Borah}
\email{anganab@umich.edu}
\affiliation{%
  \institution{University of Michigan Ann Arbor \country{United States of America}}
  \city{}
  \state{}
  }

\author{Amit Awekar}
\email{awekar@iitg.ac.in}
\affiliation{%
  \institution{Indian Institute of Technology Guwahati \country{India}}
  \city{}
  \state{}
  }
\renewcommand{\shortauthors}{Author List}

\begin{abstract} 
Large Deep Learning models are compressed and deployed for specific applications. However, current Deep Learning model compression methods do not utilize the information about the target application. As a result, the compressed models are application agnostic. Our goal is to customize the model compression process to create a compressed model that will perform better for the target application. Our method, Application Specific Compression ($ASC$), identifies and prunes components of the large Deep Learning model that are redundant specifically for the given target application. The intuition of our work is to prune the parts of the network that do not contribute significantly to updating the data representation for the given application. We have experimented with the BERT family of models for three applications: Extractive QA, Natural Language Inference, and Paraphrase Identification. We observe that customized compressed models created using $ASC$ method perform better than existing model compression methods and off-the-shelf compressed models.
\end{abstract}




\maketitle

\begin{table}[]
\begin{tabular}{|l|c|c|c|c|}
\hline
\multicolumn{1}{|c|}{\textbf{\begin{tabular}[c]{@{}c@{}}Model \\ Name\end{tabular}}} & \textbf{\begin{tabular}[c]{@{}c@{}}Layers\end{tabular}} & \textbf{\begin{tabular}[c]{@{}c@{}}Parameters\\ (Million)\end{tabular}} & \textbf{\begin{tabular}[c]{@{}c@{}}Embedding\\ Dimension\end{tabular}} & \textbf{\begin{tabular}[c]{@{}c@{}}Size \\ (MB)\end{tabular}} \\ \hline
BERT-large                                                                           & 24                                                                   & 340                                                                                  & 1024                                                         & 1340                                                                  \\ \hline
BERT-base                                                                            & 12                                                                   & 110                                                                                  & 768                                                          & 440                                                                   \\ \hline
Distil BERT                                                                          & 6                                                                    & 66                                                                                   & 768                                                          & 268                                                                   \\ \hline
BERT medium                                                                          & 8                                                                    & 41.7                                                                                 & 512                                                          & 167                                                                   \\ \hline
BERT mini                                                                            & 4                                                                    & 11.3                                                                                 & 256                                                          & 45.1                                                                  \\ \hline
Tiny BERT                                                                            & 2                                                                    & 4.4                                                                                  & 128                                                          & 17.7                                                                  \\ \hline
\end{tabular}
\caption{BERT family models used in our experiments}
\label{tab:rel}
\vspace{-1cm}
\end{table}

\section{Introduction}
Consider a Deep Learning model such as BERT and a target application such as Paraphrase Identification. It is difficult to deploy BERT on devices with a tiny amount of memory or severe power consumption restrictions. One way out is to compress it to reduce its size and energy consumption. However, we are interested in designing a model compression process that will compress BERT specifically to improve the performance of the compressed model for the paraphrase identification task. In general, how can we optimize the Deep Learning model compression process with a specific target application? 

Existing model compression methods lack the capability to optimize the compressed model for a target application. All the existing model compression methods, such as pruning, quantization, and parameter sharing, are application agnostic. They do not consider the downstream task while compressing a model. Part of a large Deep Learning model that is redundant for one application can turn out to be important for other applications. Therefore, our intuition is that an application specific compression method can create a better compressed model than existing methods.

We model the target application in terms of its training dataset. Our method, Application Specific Compression ($ASC$), analyzes changes in data representation as data flows through the given Deep Learning model from the input layer to the output. If a part of the current model does not significantly alter the representation of training data, then we consider it as redundant for the target application. We measure the cosine similarity between the embedding vectors to compute the degree of change in data representation. The higher the cosine similarity, the lower the change in data representation. $ASC$ requires only a single forward pass over the training data to identify the redundant parts of the model.

We perform experiments using the encoder-only family of models (BERT and its compressed versions) that are publicly available. Using the BERT family of models, we perform three tasks. The first task is Extractive Question Answering using the SQuAD2.0 dataset. For this task, we use exact match and F-1 score as evaluation measures. The second task is Natural Language Inference on the SNLI dataset. The third task is Paraphrase Identification using the Quora Question Pair dataset. For the last two tasks, we use accuracy as the evaluation measure. A summary of the BERT family of models is given in Table \ref{tab:rel}. We show that our method $ASC$ produces the best quality compressed models for all three tasks as compared to all the baselines. Our primary research contribution is to bring into consideration the target application while compressing a model.For complete reproducibility, all our code, data, models, and additional experimental results are publicly available on the Web\footnote{\url{https://github.com/rohitrai11/Application-Specific-Compression-of-Deep-Learning-Models}}.

\section{Related Work}
Model compression is an active field of research. The popular model compression techniques are pruning, quantization, knowledge distillation, parameter sharing, early exit, and token skipping. A general overview of these techniques can be found in \cite{10.1145/3487045}.

In \textit{pruning}, unimportant or redundant weight connections in a trained network are removed. There are different ways of pruning deep neural networks, such as pruning weights \cite{DBLP:conf/iclr/NarangDSE17}, neurons \cite{srinivas2015data}, attention heads \cite{voita-etal-2019-analyzing} and layers \cite{PEER202276} in transformer networks.

\textit{Quantization} involves compressing the models by reducing the number of bits needed to store the weights. The goal of quantization is to make the model lightweight without a significant reduction in its performance. BinaryBERT \cite{bai-etal-2021-binarybert} is a highly compressed model (24x smaller) obtained after the binarization of BERT parameters. I-BERT \cite{pmlr-v139-kim21d} performs an end-to-end integer-only BERT inference without the need for any floating-point calculation.

\textit{Knowledge Distillation} \cite{hinton2015distilling} is a process in which knowledge is condensed from a large or complex deep neural model to a simplified one. The simplified model should be able to achieve accuracy or performance similar to that of the complex model. DistilBERT \cite{sanh2019distilbert} is one of the most well-known compressed BERT models obtained by this technique. It is about 40\% smaller than BERT and retains 97\% of BERT's language understanding capabilities. TinyBERT \cite{jiao2019tinybert} is another well-known compressed BERT model that is obtained via a two-stage learning framework of knowledge distillation.

\textit{Parameter Sharing} method performs model compression by identifying weights that can be allocated the same weight value. This method has been employed in character-aware Language Models \cite{ling2015finding} in which character embeddings are learned to generate the word embeddings. Parameter sharing method has also been
employed to multiple transformer architectures \cite{lan2020albert}.

\textit{Early Exit} \cite{xin-etal-2020-deebert} speeds up the model’s inference by taking decisions at each layer, and if the condition is fulfilled, the model exits from that layer. In this way, the input does not need to go through all the layers of the model. \textit{Token Skipping} \cite{pmlr-v119-goyal20a} also escalates model inference without any size reduction. The model parameters are skipped/dropped between each layer based on their importance.

In the recent past, some compression methods have been specifically aimed at large language models (LLMs). Most of them are based on the intuition that many layers in the LLMs are redundant. Gromov et al. \cite{gromov2024unreasonable} remove a fixed number of layers from the model. In contrast, $ASC$ decides the number of layers to be pruned based on the target application. Men et al. \cite{men2024shortgpt} remove a contiguous block of layers from the model. However, $ASC$ does not impose any such restriction while removing layers. Yang et al. \cite{yang2024laco} remove layers only from the rear end of the model in an iterative manner. Whereas $ASC$ can remove layers from any part of the model in one go. Xu et al. \cite{xu2024besa} have proposed a new method named Blockwise Parameter-Efficient Sparsity Allocation (BESA). It involves pruning blocks of weights in self-attention and feed-forward networks by constructing a network pruning mask for each block. Cao et al. \cite{cao2024head} compress LLMs by sharing weights among attention weight matrices having high cosine similarity. The last two works by Xu et al. and Cao et al. are orthogonal to our $ASC$ approach of layer pruning. Both these methods can be augmented with any other model compression method that prunes layers.

\section{Experimental Set-up}
We have performed experiments with the BERT family of models. We have considered three target tasks: Extractive Question Answering (EQA), Natural Language Inference (NLI), and Paraphrase Identification (PI). We have used the following datasets to conduct our experiments.
\begin{itemize}
    \item \textbf{SQuAD} for EQA: Stanford Question Answering Dataset (SQuAD) \cite{rajpurkar2016squad} is a reading comprehension dataset. It consists of questions from a set of Wikipedia articles, where the answer to every question denotes a segment of text, or span, from the corresponding reading passage, or there may be no answer to a given question. We have used the SQuAD2.0 dataset for our experiments. It combines the 100,000 questions in SQuAD1.1 with over 50,000 unanswerable questions. The dataset is available freely on the web \footnote{\url{https://rajpurkar.github.io/SQuAD-explorer/}}.
    \item \textbf{SNLI} for NLI: The SNLI (Stanford Natural Language Inference) \cite{bowman2015large} dataset is a large-scale dataset used for training and evaluating natural language understanding models, specifically for the task of natural language inference (NLI). It consists of 570k human-written English sentence pairs, each having one of the three labels: entailment, contradiction, and neutral.
    \item \textbf{QQP} for PI: The Quora Question Pair dataset contains pairs of questions where each pair is labelled. The task is to determine whether the two questions in a pair convey the same meaning. The dataset consists of over 400k samples of training instances. For our experiments, we used 10 \% of the training set as validation, and the other 10 \% of the remaining training set was used as the test partition. The dataset was released as a part of a Kaggle competition \footnote{\url{https://www.kaggle.com/datasets/quora/question-pairs-dataset}}.
\end{itemize}

We performed all our experiments using an NVIDIA A100-SXM4 GPU with 40 GB of graphics memory. For all three tasks, we first fine-tune various models on the train partition of the dataset and report the model's performance on the test set. The number of required training epochs for each task is 3 (EQA), 5 (NLI), and 10 (PI). All models used are publicly available in the Hugging Face library. We have compared against the following four baselines:

\begin{itemize}
    \item \textbf{Baseline-1:} Large pre-trained models (BERT-large and BERT-base)
    \item \textbf{Baseline-2:} Existing model compression methods (Pruning \cite{zafrir2021prune} and 8 bit Quantization \footnote{\url{https://pytorch.org/docs/stable/quantization.html}})
    \item \textbf{Baseline-3:} Off-the-shelf compressed models (Distil-BERT, BERT-medium, and Tiny BERT)
    \item \textbf{Baseline-4:} Compressed model with random layers removed from the model
\end{itemize}

\section{Our Work}
Our method is a two-step process. First, we compute the similarity for all pairs of layers in the uncompressed model (Algorithm~\ref{algo:1}). Second, using the similarity matrix and a threshold value, we identify and prune the redundant layers in the given model (Algorithm~\ref{algo:2}). The similarity values are in the range of -1 to 1. In the case of the BERT-base, the similarity matrix will be 13X13 (one embedding layer and twelve encoder layers) with similarity values for $13 \choose 2$ pairs of layers. The similarity matrix is symmetric, with all diagonal elements equal to 1. 

Please refer to Algorithm~\ref{algo:1}. We start with the uncompressed model that is fine-tuned for the target application. It is the best-performing model that we have for the given task. We want to compress it by removing layers that do not contribute to altering the data representation. We compute the similarity using a single forward pass over the training dataset. We go through each data point and pass it through the model (the loop from lines 5 to 16). For each token in the current data point, we note the embedding vector generated by each layer (the loop from lines 6 to 15). The similarity between any two layers is the cosine similarity between the embeddings generated by those two layers (the loop from lines 9 to 14). The final similarity value is the average across all tokens in the dataset. That is why we have $N$ in the denominator on line 11.

Please refer to Algorithm~\ref{algo:2}. Consider two layers $i$ and $j$ with $i < j$. If the similarity score for this pair is high, then it indicates that when data flows from layer $i$ to layer $j$, data representation does not change significantly. In other words, we can consider all layers from $i+1$ to $j$ as redundant for the target application (line 10). We can directly connect layer $i$ with layer $j+1$. For each layer $i$ in the model, we find the farthest layer $j$ such that their similarity is above the specified threshold threshold (the loop from lines 3 to 13). Please note that we start with the highest value of $j$ and go on reducing it in the loop from lines 4 to 11. If we cannot find any such $j$ for the given value of $i$, then we move to the next layer. After we have identified all redundant layers, we remove them from the model (lines 14 and 15). Again, we fine-tuned the compressed model for the target application (line 16).

\begin{algorithm}
\caption{Similarity Matrix Creation Using Forward Pass}
\begin{algorithmic}[1]
\State \textbf{Input:} Uncompressed Deep Learning model fine-tuned for the task (fine-tuned BERT-base for our experiments), Application specific training dataset
\State \textbf{Output:} Similarity Matrix

\Comment{Perform a single forward pass of the training dataset through the fine-tuned uncompressed model to compute the similarity matrix across all layers of the model.} 

\State Initialize the similarity matrix $Sim$ with all zeros. (13X13 matrix in our experiments for the BERT-base)
\State $N$ is the total number of token in the training dataset.
\For{each data point of the training dataset}
    \For{each token of the current data point }
        \State Let current token be $Tok$
        \State Record output embedding vector of each layers (13 layers in our experiments: $O_1(Tok),\ O_2(Tok).....O_{12}(Tok)$).
        
        \Comment{Following two nested loops update the similarity matrix using the current token}
        \For{i=1 to Number of layers in the model}
            \For{j=i to Number of layers in the model}
                \State $Sim[i][j] + = (Cosine(O_i(Tok), O_i(Tok))/N)$
                \State $Sim[j][i]$=$Sim[i][j]$
            \EndFor
        \EndFor        
    \EndFor
\EndFor
\end{algorithmic}
\label{algo:1}
\end{algorithm}

\begin{algorithm}
\caption{Application Specific Compression Using Similarity Matrix}
\begin{algorithmic}[1]
\State \textbf{Input:} Similarity Matrix $Sim$ and a similarity threshold value $threshold$
\State \textbf{Output:} Fine-tuned compressed model

\Comment{Identify the redundant layers using the similarity matrix}

\For{i=1 to Number of layers in the model}
    \For{j=Number of layers in the model to i}
        \If{$Sim[i][j] \geq threshold$}
            \State Break        
        \EndIf
    \EndFor
    \If{j > i}
        \State Mark layers from $i+1$ to $j$ as redundant layers.
    \EndIf
    \State i=j+1
\EndFor

\State Prune all layers that are marked as redundant.
\State Adjust layer-wise connections after pruning.

\State Fine-tune the compressed model for the target task
\end{algorithmic}
\label{algo:2}
\end{algorithm}

\begin{figure}
\centering
\includegraphics[width=0.9\linewidth]{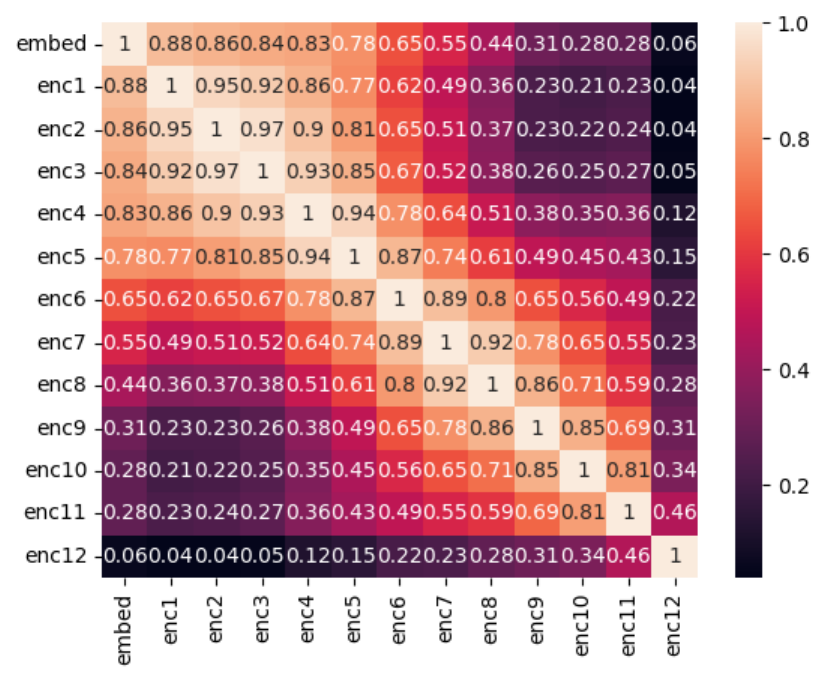}
\caption{Similarity Matrix for EQA task}
\label{fig:exqa}
\end{figure}

\begin{figure}
\centering
\includegraphics[width=0.9\linewidth]{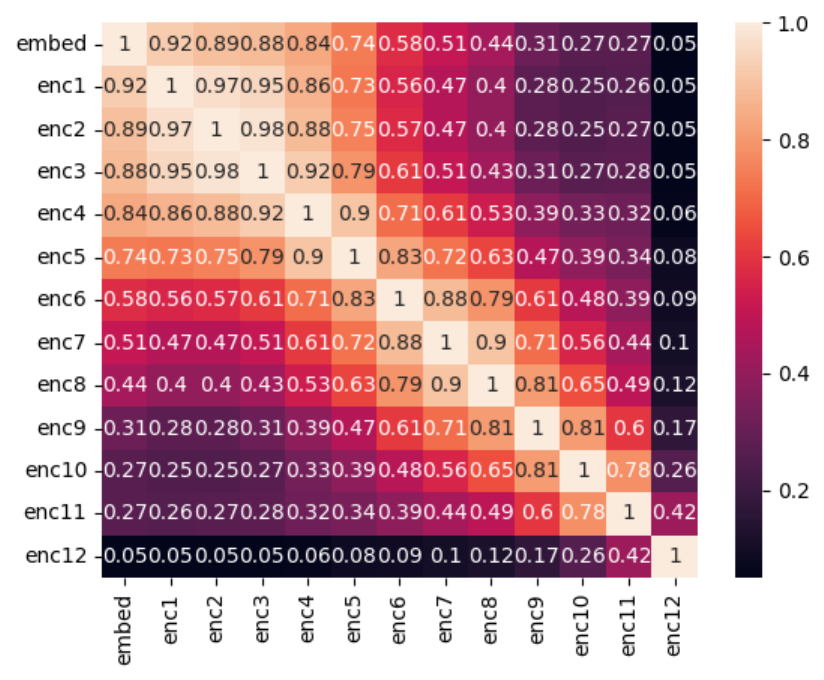}
\caption{Heat map for NLI task}
\label{fig:NLI}
\end{figure}

\begin{figure}
\centering
\includegraphics[width=0.9\linewidth]{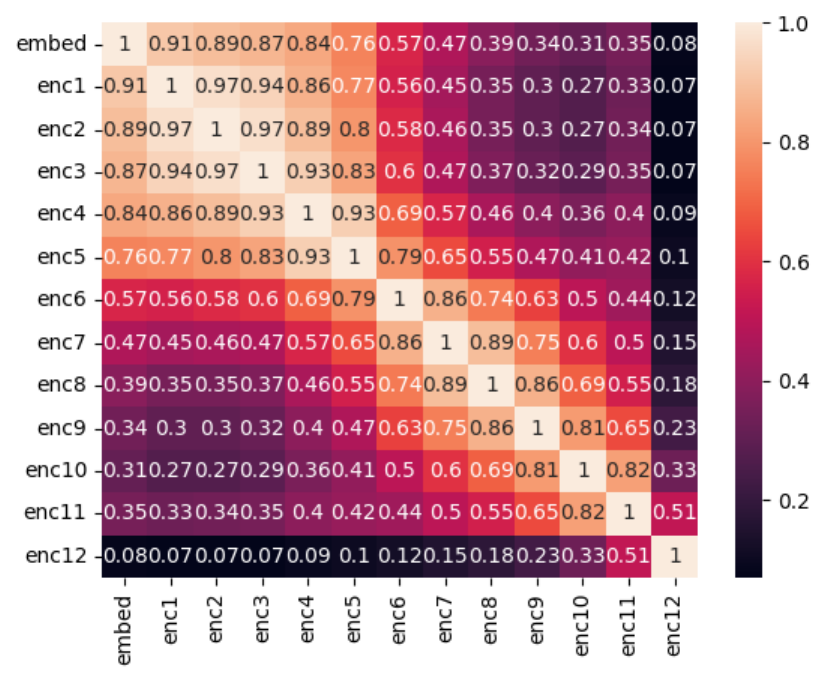}
\caption{Heat map for PI task}
\label{fig:QQP}
\end{figure}

\begin{table*}[]
\begin{tabular}{|cc|cccc|c|c|}
\hline
\multicolumn{2}{|c|}{\multirow{3}{*}{\textbf{Model}}}                                                                                                                           & \multicolumn{4}{c|}{\textbf{EQA}}                                                                                                                  & \multirow{2}{*}{\textbf{NLI}} & \multirow{2}{*}{\textbf{PI}} \\ \cline{3-6}
\multicolumn{2}{|c|}{}                                                                                                                                                          & \multicolumn{2}{c|}{\textbf{Has Answers}}                                          & \multicolumn{2}{c|}{\textbf{No Answers}}                      &                                &                               \\ \cline{3-8} 
\multicolumn{2}{|c|}{}                                                                                                                                                          & \multicolumn{1}{c|}{\textbf{Exact Match}} & \multicolumn{1}{c|}{\textbf{F1 score}} & \multicolumn{1}{c|}{\textbf{Exact Match}} & \textbf{F1 score} & \textbf{Accuracy}              & \textbf{Accuracy}             \\ \hline
\multicolumn{1}{|c|}{\multirow{2}{*}{\textbf{Baseline-1}}}                                                  & BERT-large                                                        & \multicolumn{1}{c|}{76.31579}             & \multicolumn{1}{c|}{83.43005}          & \multicolumn{1}{c|}{78.04878}             & 78.04878          & 90.52                          & 91.06                         \\ \cline{2-8} 
\multicolumn{1}{|c|}{}                                                                                      & BERT-base                                                         & \multicolumn{1}{c|}{73.49865}             & \multicolumn{1}{c|}{80.31598}          & \multicolumn{1}{c|}{71.92599}             & 71.92599          & 90.1                           & 90.86                         \\ \hline
\multicolumn{1}{|c|}{\multirow{2}{*}{\textbf{Baseline-2}}}                                                  & Pruned BERT                                                       & \multicolumn{1}{c|}{66.4305}              & \multicolumn{1}{c|}{72.52775}          & \multicolumn{1}{c|}{69.26829}             & 69.26829          & 88.62                          & 90.28                         \\ \cline{2-8} 
\multicolumn{1}{|c|}{}                                                                                      & Quantized BERT                                                    & \multicolumn{1}{c|}{55.85358}             & \multicolumn{1}{c|}{63.00107}          & \multicolumn{1}{c|}{79.44491}             & 79.44491          & 87.53                          & 89.72                         \\ \hline
\multicolumn{1}{|c|}{\multirow{4}{*}{\textbf{Baseline-3}}}                                                  & Distil-BERT                                                       & \multicolumn{1}{c|}{66.10999}             & \multicolumn{1}{c|}{72.84249}          & \multicolumn{1}{c|}{61.49706}             & 61.49706          & 87.9                           & 89.91                         \\ \cline{2-8} 
\multicolumn{1}{|c|}{}                                                                                      & BERT-medium                                                       & \multicolumn{1}{c|}{67.05466}             & \multicolumn{1}{c|}{73.52532}          & \multicolumn{1}{c|}{65.13036}             & 65.13036          & 88.18                          & 89.9                          \\ \cline{2-8} 
\multicolumn{1}{|c|}{}                                                                                      & BERT-mini                                                         & \multicolumn{1}{c|}{44.58502}             & \multicolumn{1}{c|}{51.38589}          & \multicolumn{1}{c|}{64.03701}             & 64.03701          & 85.13                          & 88.41                         \\ \cline{2-8} 
\multicolumn{1}{|c|}{}                                                                                      & Tiny BERT                                                         & \multicolumn{1}{c|}{14.7942}              & \multicolumn{1}{c|}{17.71541}          & \multicolumn{1}{c|}{83.1455}              & 83.1455           & 79.9                           & 84.19                         \\ \hline
\multicolumn{1}{|c|}{\textbf{Baseline-4}}                                                                   & \begin{tabular}[c]{@{}c@{}}Random 6 layers\\ removed\end{tabular} & \multicolumn{1}{c|}{60.9143}              & \multicolumn{1}{c|}{69.37958}          & \multicolumn{1}{c|}{55.08831}             & 55.08831          & 88.45                          & 89.4                          \\ \hline
\multicolumn{1}{|c|}{\multirow{3}{*}{\textbf{\begin{tabular}[c]{@{}c@{}}ASC\end{tabular}}}} & Threshold = 90\%                                                  & \multicolumn{1}{c|}{\textbf{67.66194}}    & \multicolumn{1}{c|}{\textbf{75.01801}} & \multicolumn{1}{c|}{\textbf{69.67199}}    & \textbf{69.67199} & \textbf{90.05}                 & \textbf{90.76}                \\ \cline{2-8} 
\multicolumn{1}{|c|}{}                                                                                      & Threshold = 85\%                                                  & \multicolumn{1}{c|}{65.23279}             & \multicolumn{1}{c|}{72.72148}          & \multicolumn{1}{c|}{68.51135}             & 68.51135          & 89.12                          & 90.04                         \\ \cline{2-8} 
\multicolumn{1}{|c|}{}                                                                                      & Threshold = 80\%                                                  & \multicolumn{1}{c|}{55.60054}             & \multicolumn{1}{c|}{64.0874}           & \multicolumn{1}{c|}{62.20353}             & 62.20353          & 88.01                          & 89.42                         \\ \hline
\end{tabular}
\caption{Performance comparison of model compression methods for three tasks}
\label{tab:1}
\vspace{-4mm}
\end{table*}

\section{Results and Discussion}
Please refer to Figure~\ref{fig:exqa}. It shows the similarity matrix for the EQA task using the SQuAD2.0 dataset over the fine-tuned BERT-base model. We can observe that similarity scores for initial layers are comparatively higher than for later layers. Initial layers generally capture more general and surface-level features of the input text, such as basic syntactic information and word-level similarities. In contrast, deeper layers capture more abstract and task-specific features. The similarity between the outputs of consecutive layers is high. This happens because word representations are refined incrementally by learning from the previous layer. We can observe a similar pattern for NLI (Figure~\ref{fig:NLI}) and PI (Figure\ref{fig:QQP}) tasks.

We use these similarity matrices to determine redundant layers for each task. We have experimented with three similarity threshold values: 0.9. 0.85, and 0.8. Using the three thresholds, we have created three compressed models using our method $ASC$. For the EQA task, when the threshold is 80\%, we are able to delete seven encoder layers: 1,2,3,4,6,8 and 10. On increasing the threshold to 85\%, we are able to delete only five encoder layers: 1,2,4,6 and 8. Finally, when the threshold is set to 90\%, we are able to delete only four encoder layers: 2,3,5 and 8. For the NLI task, when the threshold is 80\%, we are able to delete seven encoder layers: 1,2,3,4,6,8 and 10. For the 85\% threshold, we can prune five encoder layers: 1,2,4,5 and 7. And for the 90\% threshold, we delete only four encoder layers: 1,3,5, and 8. For the PI task, when the threshold is 80\%, we are able to delete seven encoder layers: 1,2,3,4,7,9 and 11. For the 85\% threshold, we delete six encoder layers: 1,2,3,5,7 and 9. With the 90\% threshold, we delete only three encoder layers: 1,3, and 5. 

Please refer to Table~\ref{tab:1}. It shows the performance comparison of $ASC$ with various baselines. The performance of both uncompressed models in Baseline-1 is highest for all three tasks. It is expected because these models have larger sizes than compressed models. It will not be possible for any compressed model to perform better than the original uncompressed model. However, $ASC$ with a 90\% similarity threshold is able to beat all compressed model baselines. Performance gaps between $ASC$ (90\% threshold) and BERT-base are quite narrow. It indicates that $ASC$ is good at identifying redundant parts of the uncompressed model.

\section{Conclusion and Future Work}
We introduced Application Specific Compression method for compressing Deep Learning models with a specific target application. Our method effectively identifies redundant parts of the model by performing a single forward pass using the training dataset for the target application. Our method performs better than all baseline compressed models. Currently, we identify redundant parts of the network in a single shot. Our $ASC$ method can be further improved by introducing an iterative version of application specific pruning.

\bibliographystyle{ACM-Reference-Format}
\bibliography{sample-base}

\appendix

\end{document}